\documentclass[11pt, a4paper]{fair}
\usepackage[numbers,sort&compress]{natbib}

\setheadertext{Towards Concreteness Paved Contrastive Negative Mining for Compositional Understanding}

\renewcommand{\today}{}

\usepackage[utf8]{inputenc}
\usepackage[T1]{fontenc}
\usepackage{hyperref}
\usepackage{url}
\usepackage{tabularx}
\usepackage{makecell}
\usepackage{array,ragged2e,booktabs}
\newcolumntype{P}[1]{>{\RaggedRight\arraybackslash}p{#1}}
\usepackage{longtable}
\usepackage{amsfonts}
\usepackage{amsthm}

\usepackage{nicefrac}
\usepackage{microtype}
\usepackage{xcolor}
\usepackage{xspace}
\newcommand{\eg}{\textit{e.g.}\xspace}
\usepackage{algorithm}
\usepackage{algorithmicx}
\usepackage{algpseudocode}
\usepackage{bxcoloremoji}

\definecolor{darkblue}{rgb}{0, 0, 0.5}
\hypersetup{colorlinks=true, citecolor=darkblue, linkcolor=darkblue, urlcolor=magenta}

\usepackage{xurl}

\usepackage{soul}
\usepackage{amsmath}
\usepackage{subcaption}
\usepackage{graphicx}
\usepackage{mathtools}
\usepackage{pifont}
\usepackage{tocloft}
\usepackage{lipsum}
\usepackage{colortbl}
\usepackage{wrapfig}
\usepackage{multirow}
\usepackage{enumitem}
\usepackage{cleveref}
\usepackage{listings}
\usepackage{fontawesome5} 
\usepackage{caption}
\usepackage{afterpage}
\usepackage[edges]{forest}
\usepackage[normalem]{ulem}
\usepackage{CJKutf8}
\usepackage{bbding}
\usepackage[tikz]{bclogo}
\usepackage[framemethod=tikz]{mdframed}

\definecolor{darkblue}{rgb}{0, 0, 0.5}
\hypersetup{colorlinks=true, citecolor=darkblue, linkcolor=darkblue, urlcolor=darkblue}
\newcommand{\datapipe}{ConcretePlant}
\newcommand{\model}{Slipform}

\newcommand{\wrt}{\textit{w.r.t.}}
\newcommand{\etal}{\textit{et al.}}

\definecolor{HeaderBlue}{RGB}{234,242,255}

\definecolor{headerblue}{HTML}{2596be}

\mdfdefinestyle{mystyle}{%
  rightline=true,
  innerleftmargin=10,
  innerrightmargin=10,
  outerlinewidth=3pt,
  topline=false,
  rightline=true,
  bottomline=false,
  skipabove=\topsep,
  skipbelow=\topsep
}

\newtcolorbox{myboxic}[1][]{
  breakable,
  title=#1,
  colback=red!5,
  colbacktitle=red!5,
  coltitle=black,
  fonttitle=\bfseries,
  bottomrule=0pt,
  toprule=0pt,
  leftrule=2pt,
  rightrule=2pt,
  titlerule=0pt,
  arc=0pt,
  outer arc=0pt,
  colframe=red,
}

\newtcolorbox{myboxi}[1][]{
  breakable,
  title=#1,
  colback=yellow!6!white,
  colbacktitle=yellow!6!white,
  coltitle=black,
  fonttitle=\bfseries,
  bottomrule=0pt,
  toprule=0pt,
  leftrule=2pt,
  rightrule=2pt,
  titlerule=0pt,
  arc=0pt,
  outer arc=0pt,
  colframe=orange!50!yellow,
}

\newtcolorbox{myboxis}[1][]{
  breakable,
  title=#1,
  colback=Indigo!5,
  colbacktitle=Indigo!5,
  coltitle=black,
  fonttitle=\bfseries,
  bottomrule=0pt,
  toprule=0pt,
  leftrule=2pt,
  rightrule=2pt,
  titlerule=0pt,
  arc=0pt,
  outer arc=0pt,
  colframe=Indigo,
}

\usepackage{epigraph}

\newtcolorbox{myboxii}[1][]{
  breakable,
  freelance,
  title=#1,
  colback=white,
  colbacktitle=white,
  coltitle=black,
  fonttitle=\bfseries,
  bottomrule=0pt,
  boxrule=0pt,
  colframe=white,
  overlay unbroken and first={
  \draw[red!75!black,line width=3pt]
    ([xshift=5pt]frame.north west) -- 
    (frame.north west) -- 
    (frame.south west);
  \draw[red!75!black,line width=3pt]
    ([xshift=-5pt]frame.north east) -- 
    (frame.north east) -- 
    (frame.south east);
  },
  overlay unbroken app={
  \draw[red!75!black,line width=3pt,line cap=rect]
    (frame.south west) -- 
    ([xshift=5pt]frame.south west);
  \draw[red!75!black,line width=3pt,line cap=rect]
    (frame.south east) -- 
    ([xshift=-5pt]frame.south east);
  },
  overlay middle and last={
  \draw[red!75!black,line width=3pt]
    (frame.north west) -- 
    (frame.south west);
  \draw[red!75!black,line width=3pt]
    (frame.north east) -- 
    (frame.south east);
  },
  overlay last app={
  \draw[red!75!black,line width=3pt,line cap=rect]
    (frame.south west) --
    ([xshift=5pt]frame.south west);
  \draw[red!75!black,line width=3pt,line cap=rect]
    (frame.south east) --
    ([xshift=-5pt]frame.south east);
  },
}

\definecolor{myblue}{rgb}{0.9, 0.1, 0.94}
\definecolor{mygreen}{rgb}{0.64, 0.56, 0.88}
\definecolor{myyellow}{rgb}{0.68, 0.6, 0.1}
\definecolor{fancygreen}{rgb}{0.33, 0.68, 0.20}
\definecolor{salmon}{rgb}{0.94, 0.52, 0.49}
\definecolor{tablegreen}{rgb}{0.82, 0.94, 0.75}
\definecolor{tableblue}{rgb}{0.81, 0.90, 0.94}
\definecolor{tablered}{rgb}{0.97, 0.85, 0.85}
\definecolor{tableorange}{rgb}{0.96, 0.85, 0.81}

\definecolor{myorange}{rgb}{1.0, 0.49, 0.0}	
\definecolor{tlgreen}{rgb}{0.33, 0.68, 0.20}

\newenvironment{itemize*}%
 {\leftmargini=10pt\begin{itemize}%
  \setlength{\itemsep}{0pt}%
  \setlength{\parskip}{0pt}%
  }%
 {\end{itemize}}
\newenvironment{enumerate*}%
 {\begin{enumerate}%
  \setlength{\itemsep}{0pt}%
  \setlength{\parskip}{0pt}}%
 {\end{enumerate}}

\definecolor{sptColor}{HTML}{E60000} 
\definecolor{rlColor}{HTML}{F07A00}      
\definecolor{distillColor}{HTML}{FFE000}         
\definecolor{aignmentColor}{HTML}{69b847}   
\definecolor{benchmarkColor}{HTML}{0095FF}

\tikzset{%
    every node/.style={font=\tiny},
    parent/.style =          {align=center,text width=2cm,rounded corners=3pt, line width=0.3mm, fill=gray!10,draw=gray!80},
    child/.style =           {align=center,text width=2.0cm,rounded corners=3pt, fill=blue!10,draw=blue!80,line width=0.3mm},
    grandchild/.style =      {align=center,text width=2cm,rounded corners=3pt},
    greatgrandchild/.style = {align=center,text width=1.5cm,rounded corners=3pt},
    greatgrandchild2/.style = {align=center,text width=1.5cm,rounded corners=3pt},    
    referenceblock/.style =  {align=center,text width=1.5cm,rounded corners=2pt},
    pretrain/.style =           {align=center,text width=2.0cm,rounded corners=3pt, fill=aignmentColor!10,draw=aignmentColor!80,line width=0.3mm},   
    pretrain_work/.style =           {align=center, text width=8.5cm,rounded corners=3pt, fill=aignmentColor!10,draw=aignmentColor!0,line width=0.3mm},  
    template/.style =           {align=center,text width=2.0cm,rounded corners=3pt, fill=rlColor!10,draw=rlColor!80,line width=0.3mm},   
    template_work/.style =           {align=center,text width=8.5cm,rounded corners=3pt, fill=rlColor!10,draw=rlColor!0,line width=0.3mm},    
    answer/.style =           {align=center,text width=2.0cm,rounded corners=3pt, fill= sptColor!10,draw= sptColor!80,line width=0.3mm},   
    answer_work/.style =           {align=center,text width=8.5cm,rounded corners=3pt, fill= sptColor!10,draw= sptColor!0,line width=0.3mm},      
    multiple/.style =           {align=center,text width=2.0cm,rounded corners=3pt, fill= distillColor!10,draw= distillColor!80,line width=0.3mm},   
    multiple_work/.style =           {align=center,text width=8.5cm,rounded corners=3pt, fill= distillColor!10,draw= distillColor!0,line width=0.3mm},        
    tuning/.style =           {align=center,text width=2.0cm,rounded corners=3pt, fill= benchmarkColor!10,draw= benchmarkColor!80,line width=0.3mm},   
    tuning_work/.style =           {align=center,text width=8.5cm,rounded corners=3pt, fill= benchmarkColor!10,draw= benchmarkColor!0,line width=0.3mm},          
}

\usepackage{tikz}

\newcommand{\lstbg}[3][0pt]{{\fboxsep#1\colorbox{#2}{\strut #3}}}

\lstdefinelanguage{diff}{
  basicstyle=\ttfamily\small,
  morecomment=[f][\lstbg{red!20}]-,
  morecomment=[f][\lstbg{green!20}]+,
}
\lstdefinelanguage{diffpython}{
  language=diff,
  morekeywords={def, if, else, for, while, return, import, from, as, class, with, try, except, finally, raise, lambda, and, or, not, in, is, None, True, False},
  morecomment=[l]{\#},
  morestring=[b]",
  morestring=[b]',
}

\definecolor{darkgreen}{RGB}{50,100,0}
\definecolor{darkred}{RGB}{200, 0, 0}
\definecolor{lightblue}{RGB}{220,235,250}

\tcbset{
  takeawaysbox/.style={
    title=Takeaways,
    colback=lightblue!80,
    colframe=black,
    fonttitle=\bfseries\small,
    coltitle=white,
    colbacktitle=black,
    enhanced,
    attach boxed title to top left={xshift=2.5mm,yshift=-2.5mm},
    boxed title style={rounded corners, size=small, colframe=black, colback=black},
    width=\linewidth,
    arc=3.5mm
  }
}

\definecolor{darkgreen}{RGB}{50,100,0}
\definecolor{darkred}{RGB}{200, 0, 0}

\NewDocumentCommand{\kaiyan}
{ mO{} }{\textcolor{purple}{\textsuperscript{\textit{kaiyan}}\textsf{\textbf{\small[#1]}}}}
\NewDocumentCommand{\yuxin}
{ mO{} }{\textcolor{cyan}{\textsuperscript{\textit{yuxin}}\textsf{\textbf{\small[#1]}}}}
\NewDocumentCommand{\bx}
{ mO{} }{\textcolor{green}{\textsuperscript{\textit{bx}}\textsf{\textbf{\small[#1]}}}}
\NewDocumentCommand{\at}
{ mO{} }{\textcolor{red}{\textsuperscript{\textit{AT}}\textsf{\textbf{\small[#1]}}}}
\NewDocumentCommand{\re}
{ mO{} }{\textcolor{blue}{\textsuperscript{\textit{RE}}\textsf{\textbf{\small[#1]}}}}
\NewDocumentCommand{\ybsun}
{ mO{} }{\textcolor{magenta}{\textsuperscript{\textit{youbang}}\textsf{\textbf{\small[#1]}}}}
\NewDocumentCommand{\runze}
{ mO{} }{\textcolor{orange}{\textsuperscript{\textit{runze}}\textsf{\textbf{\small[#1]}}}}

\definecolor{darkgreen}{RGB}{0,100,0} 
\NewDocumentCommand{\add}
{ mO{} }{\textcolor{darkgreen}{\textsuperscript{\textit{Maybe Consider Discuss}}\textsf{\textbf{[#1]}}}}

\setlist[itemize]{leftmargin=20pt}

\definecolor{hidden-blue}{RGB}{194,232,247}
\definecolor{hidden-black}{RGB}{20,68,106}

\usepackage{tabularx}
\usepackage{xstring}
\usepackage[export]{adjustbox}

\definecolor{yes}{HTML}{C6EFCE}      %
\definecolor{no}{HTML}{FFC7CE}       %
\definecolor{partial}{HTML}{FFEB9C}  %
\definecolor{external}{HTML}{D9E1F2} %
\definecolor{hdr}{HTML}{F2F2F2}

\usepackage{amssymb}
\newcommand{\cmark}{\textcolor{darkgreen}{\boldmath$\checkmark$}}
\newcommand{\xmark}{\textcolor{darkred}{\boldmath$\times$}}

\newcommand{\cellstatus}[1]{%
  \begingroup
  \StrTrim{#1}[\statusval]%
  \IfStrEq{\statusval}{Yes}{\cellcolor{yes}\cmark}{}%
  \IfStrEq{\statusval}{No}{\cellcolor{no}\xmark}{}%
  \IfBeginWith{\statusval}{Yes (}{\cellcolor{yes}\cmark~\textit{\statusval\unskip}}{}%
  \IfStrEq{\statusval}{Partial}{\cellcolor{partial}\textbf{Partial}}{}%
  \IfStrEq{\statusval}{External}{\cellcolor{external}\textbf{External}}{}%
  \endgroup
}

\newif\ifmaintitle
\maintitletrue
\title{Concrete Jungle: Towards Concreteness Paved Contrastive Negative Mining for Compositional Understanding}

\author{
    Eun Woo Im, Dhruv Madhwal, Vivek Gupta \\
    Arizona State University
}

\abstract{\begin{abstract}
Vision-Language Models demonstrate remarkable capabilities but often struggle with compositional reasoning, exhibiting vulnerabilities regarding word order and attribute binding. This limitation arises from a scarcity of informative samples needed to differentiate subtle semantic variations during contrastive pretraining. Although hard negative mining offers a promising remedy, existing methods lack explicit mechanisms to dictate which linguistic elements undergo modification. Instead of engineering generative architectures, this study establishes lexical concreteness as a fundamental determinant of negative sample efficacy. Modifying highly concrete terms generates more pronounced structural and visual discrepancies, providing a substantially stronger learning signal. Leveraging this principle, ConcretePlant is proposed to systematically isolate and manipulate perceptually grounded concepts. Analyses of the InfoNCE further reveals a severe gradient imbalance, where easily distinguishable pairs disproportionately overwhelm the optimization process and restrict the bandwidth available for nuanced learning. To resolve this degradation, the Cement loss is formulated utilizing a margin-based approach. By correlating psycholinguistic scores with sample difficulty, this objective dynamically calibrates the penalization applied to individual training pairs. Comprehensive evaluations substantiate these theoretical claims. The integrated framework, designated as Slipform, achieves state-of-the-art accuracy across diverse compositional evaluation benchmarks, general cross-modal retrieval, single and multi label linear probing.
\end{abstract}}

\resource{\faEnvelope\ \textbf{Contact: }}{
  \{eunwooim, dmadhwal, vgupt140\}@asu.edu \\
  \textbf{Project page: }\ 
  \url{https://eunwooim.github.io/concrete-jungle}
}

\begin{document}

\maketitle
\maintitlefalse 

\section{Introduction}
Vision-Language Models (VLMs) have achieved remarkable performance across diverse multimodal understanding benchmarks~\citep{radford2021learning, zhai2023sigmoid} and have been well-established as representation modules in multimodal learning, retrieval, and generation.
Throughout contrastive pretraining on web-scale image-caption datasets~\citep{sharma2018conceptual, srinivasan2021wit, schuhmann2022laion, gadre2023datacomp, chuang2025meta, wang2025scaling} within a shared embedding space, VLMs learn holistic and robust cross-modal representations.
Despite these capabilities, VLMs are prone to biases stemming from limitations of the contrastive pretraining mechanism.
These include failures to understand negations~\citep{alhamoud2025vision}, object counting~\citep{parcalabescu2022valse}, spatial relations~\citep{zhao2022vl} and accurate entity associations~\citep{alabdulmohsin2024clip}.

One of the most significant limitations is bag-of-words-like behavior, which manifests as an inability to understand word order~\citep{thrush2022winoground, koishigarina2025clip} or attribute binding~\citep{robinson2020contrastive}.
This limitation is primarily a byproduct of the contrastive objective.
In the standard multimodal contrastive pretraining, training batches are curated by randomly sampled data points from the training dataset corpus.
However, this random sampling often fails to provide effective negative samples to differentiate compositional semantics for each data point.
Therefore, VLMs tend to match global summaries rather than the grammatical structure of the caption or the fine-grained details of the scene. As a result, they may recognize the presence of objects while overlooking the underlying logic and relational bindings.

Recent studies demonstrate that \textit{hard negative mining} and the inclusion of these mined samples in the training batch can significantly mitigate this problem and improve compositional understanding~\citep{zhai2023sigmoid, zhang2024contrasting}.
Hard negatives are defined as misaligned image-caption pairs nevertheless overlaps the semantics significantly with the corresponding positive samples.
Training on these pairs forces the model to differentiate between subtly distinct inputs.
Hard negative generation techniques have evolved from simple text swapping using Part-Of-Speech (POS) tagging~\citep{honnibal2017spacy, yuksekgonul2022and} and the application of handcrafted heuristics for spatial relationship augmentation~\citep{zhang2024countercurate}, to the utilization of Large Language Models (LLMs)~\citep{fan2023improving} and image editing or generative models~\citep{patel2024tripletclip}.

However, hard negative mining fundamentally samples from a \textit{one-to-many} distribution of plausible compositional variations.
State-of-the-art methods typically design the hard negative generator by prompting LLMs to perturb a single keyword within the caption~\citep{fan2023improving, patel2024tripletclip}.
Consequently, the generation process may vary and can introduce quality difference depending on the keyword selected by the LLM.
This brittleness raises a research question: \textit{How can we maximize the quality of the hard negative samples and improve the effectiveness of those synthetic data?}
Rather than designing a better generator structure, we shift the focus to identifying the \textit{data quality factor} and provide a solution to better utilize the generated data.

This study shows that the \textit{lexical concreteness} of a selected keyword is directly related to the structural discrepancy of the resulting hard negatives, and introduces a method to control this factor.
Lexical concreteness~\citep{turney2011literal, paivio2013imagery, hessel2018quantifying} how strongly a concept is associated with \textit{perceptible} entities or experiences.
Focusing on highly concrete entities encourages the generative model to modify a localized physical concept rather than to make an ambiguous global transformation.
We then compare the datasets generated with controlled levels of concreteness to verify the central hypothesis.

Furthermore, we identify that the gradient imbalance increases substantially with batch size and show that it remains vulnerable even in small-batch settings.
To mitigate this problem, a simple margin regularization is proposed with theoretical and empirical demonstration of its controllability.
Analysis shows that logit gaps are correlated with keyword concreteness, and that the data generation process introduces a skewed prior over concreteness scores.
Based on these observations, the margin is modeled with a Fermi-Dirac distribution to adaptively determine its value and balance gradient magnitudes during the training process.

To this end, we introduce \textbf{Concrete Jungle}, a concreteness-aware pretraining method that includes the following contributions:
(1) We identify the critical role of lexical concreteness in hard negative mining. Building on this, we propose \textbf{ConcretePlant}, an automated hard negative generation method that explicitly targets concrete concepts to curate more effective hard negatives.
(2) We reveal the existence of gradient magnitude polarity even in small-batch settings. To address this, we introduce \textbf{Cement loss}, a simple yet effective loss function that utilizes an adaptive margin to regulate gradient magnitudes.
(3) Through extensive analysis and experiments, concreteness is shown to be a statistically significant factor in training data, and the proposed method is demonstrated to achieve superior performance across benchmarks.

\section{Maximizing Hard Negative Efficacy}
\subsection{Concreteness Matters in Hard Negative Mining}
\begin{figure*}[t]
    \centering
    \includegraphics[width=\linewidth]{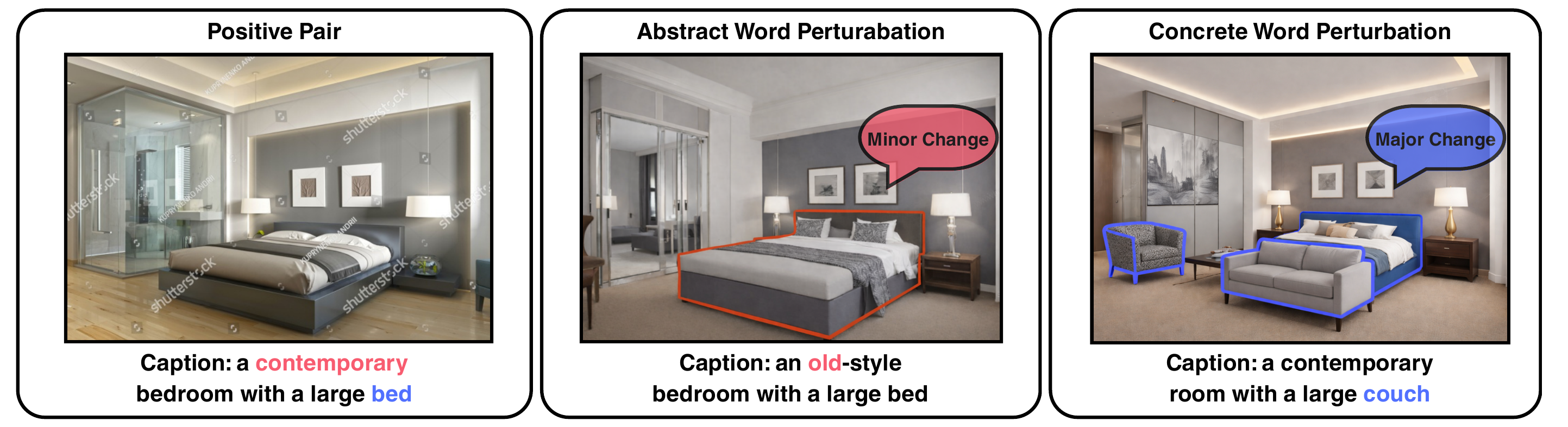}
    \vspace{-1em}
    \caption{
    Examples of our main hypothesis.
    Perturbing abstract keywords yields minor visual changes, whereas concrete words produce larger structural changes.
    }
    \vspace{-1em}
    \label{fig:teaser}
\end{figure*}

Suppose that an anchor image-caption pair  $(v_i, t_i)$ is sampled from the data corpus $\mathcal{D}$.
A standard hard negative generation pipeline typically prompts a LLM to identify a semantic keyword in the given caption $t_i$ to generate a textual perturbation $t_i'$.
Subsequently, a hard negative image $v_i'$ is obtained by feeding $v_i$ and $t_i'$ to an image generative model~\citep{patel2024tripletclip}.
As shown in Fig.~\ref{fig:teaser}, visual perceptibility of the resulting samples depends on how perceptible the keywords are.
Perturbing perceptible keywords results in structural difference, while modifying abstract stylistic descriptor yields minimal visual distinction.
We hypothesize that this indistinguishable variation between the $v_i$ and $v_i'$ fails to provide an effective learning signal.

This observation establishes the core hypothesis of the proposed methodology: psycholinguistic concreteness score $c_i$~\citep{brysbaert2014concreteness} of a selected text token is strongly dependent on the physical difference of the resulting hard negative image.
Maximizing this metric during the keyword selection process systematically guarantees the visual semantic perturbation for cross-modal representation learning.
Based on this idea, the following section explains a method for concreteness-aware hard negative mining.

\subsection{\datapipe: Automating Concrete Negative Generation}
\label{sec:data-pipeline}

\begin{figure*}[t]
    \centering
    \includegraphics[width=\textwidth]{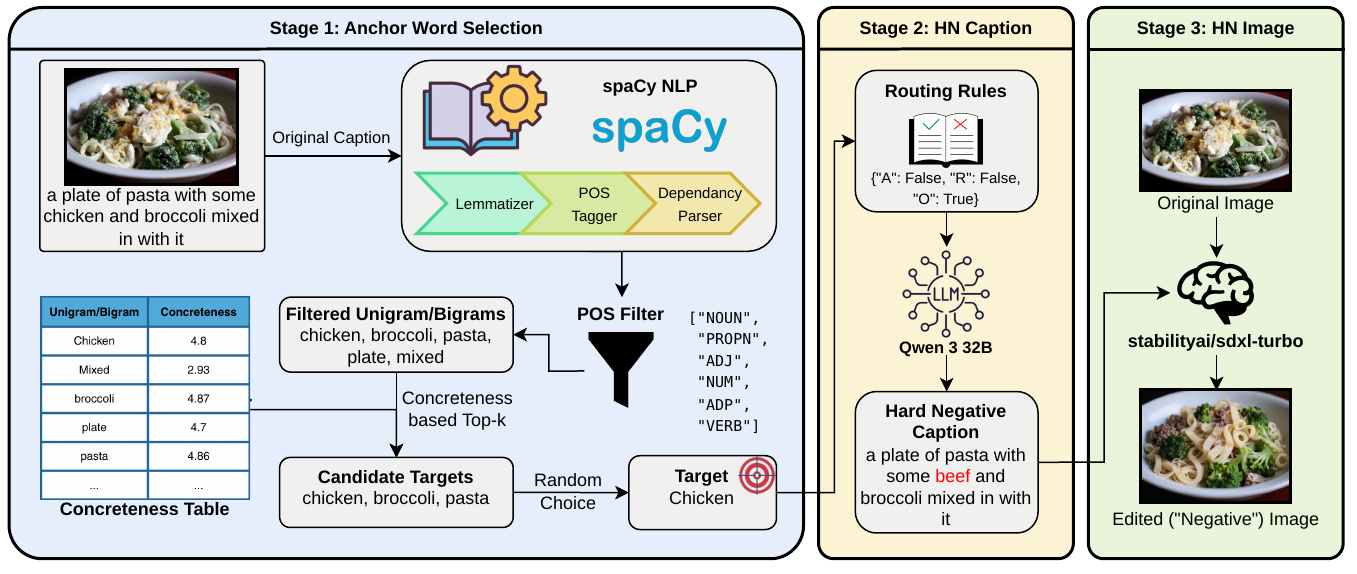}
    \vspace{-1em}
    \caption{Overview of ConcretePlant. The pipeline selects a concrete target word from the caption, generates a hard negative caption with Qwen3-32B, and edits the image with SDXL-Turbo to produce a hard negative pair.}
    \vspace{-1em}
    \label{fig:concreteplant}
\end{figure*}
\paragraph{Concreteness-Aware Keyword Sampling}
Given an image-caption sample $(v_i,t_i)$ in source data corpus $\mathcal{D}$, the longest caption is selected to ensure semantic richness~\citep{doveh2023dense, lai2024revisit} if $\mathcal{D}$ offers multiple caption pairs.
$t_i$ is decomposed into $n$ semantic keywords $t_i=\{t_i^{k}\}_{k=1}^{n}$ with SpaCy~\citep{honnibal2017spacy}, and corresponding concreteness scores $c_i^{k}=\text{lookup}(t_i^k,\mathcal{R})$ are looked up from concreteness ratings database $\mathcal{R}$~\citep{brysbaert2014concreteness}.
Top-K sampling is first performed to ensure the preference towards higher concreteness scores, then the lexical concreteness score $c_i=\text{sample}(\text{top-K}(c_i^{1:n}))$ is defined by simple rule based sampling to balance \texttt{attribute}, \texttt{object}, \texttt{relation} (ARO) compositional categories.
Additional details are referred to Appendix~\ref{appendix:method-detail}.

\paragraph{Hard Negative Mining}
Data generation employs two generative models: Qwen3-32B~\citep{yang2025qwen3} as the hard negative captioner $f_\text{cap}$, and SDXL-Turbo~\citep{sauer2024adversarial} for hard negative image generator $f_\text{gen}$.
Given an anchor sample $(v_i,t_i,c_i)$, the LLM is instructed to perturb the selected keyword while minimizing the residual of the remainders $t_i'=f_\text{cap}(t_i,c_i)$.
Inspired from recent prompting methods~\citep{ali2025harnessing, im2025self}, different in-context examples are provided to $f_\text{cap}$ by compositional types to reflect the distinct linguistic roles of each AROs.
Hard negative image $v_i'=f_\text{gen}(v_i,t_i')$ is configured with high strength to avoid style or semantic transfer, and low inference steps to minimize computational budget.
Fig.~\ref{fig:concreteplant} visualize the generation pipeline, datasets generated by the lexical concreteness-controlled ConcretePlant is referred to ConcreteBatch.

\subsection{Learning \model~from Cement Loss}
\label{sec:adaptive-margin}
The standard practice of hard negative training optimizes the VLM $f_{\theta}$ with InfoNCE~\citep{oord2018representation}, by explicitly appending hard negative counterparts $(v_i',t_i')$ of each anchor samples $(v_i,t_i)$ in the training batch, thereby doubling the total batch size as:
\begin{equation}
    L_{v \to t} = -\frac{1}{2N}\sum_{i=1}^{2N}\left[ \log \frac{\exp(s_{i,i})}{\sum_{j=1}^{2N} \exp (s_{i,j})} \right] = - \mathbb{E}_{i \sim [1,2N]}\left[ \log p_{i,i}\right].
\end{equation}
$s_{i,j}=\cos(f_{\theta}(t_i),f_{\theta}(v_j))/\tau$ and $p_{i,j}=\text{softmax}([s_{i,1:2N}])[j]$ indicate cross-modal feature cosine similarities scaled with temperature $\tau$ and anchor recall, respectively.
For notational convenience, $i'$ denotes the index of hard negative counterpart of $i$-th sample within the batch, and the other $2N-2$ samples are considered as easy negatives.
\begin{figure*}[t]
    \centering
    \includegraphics[width=\linewidth]{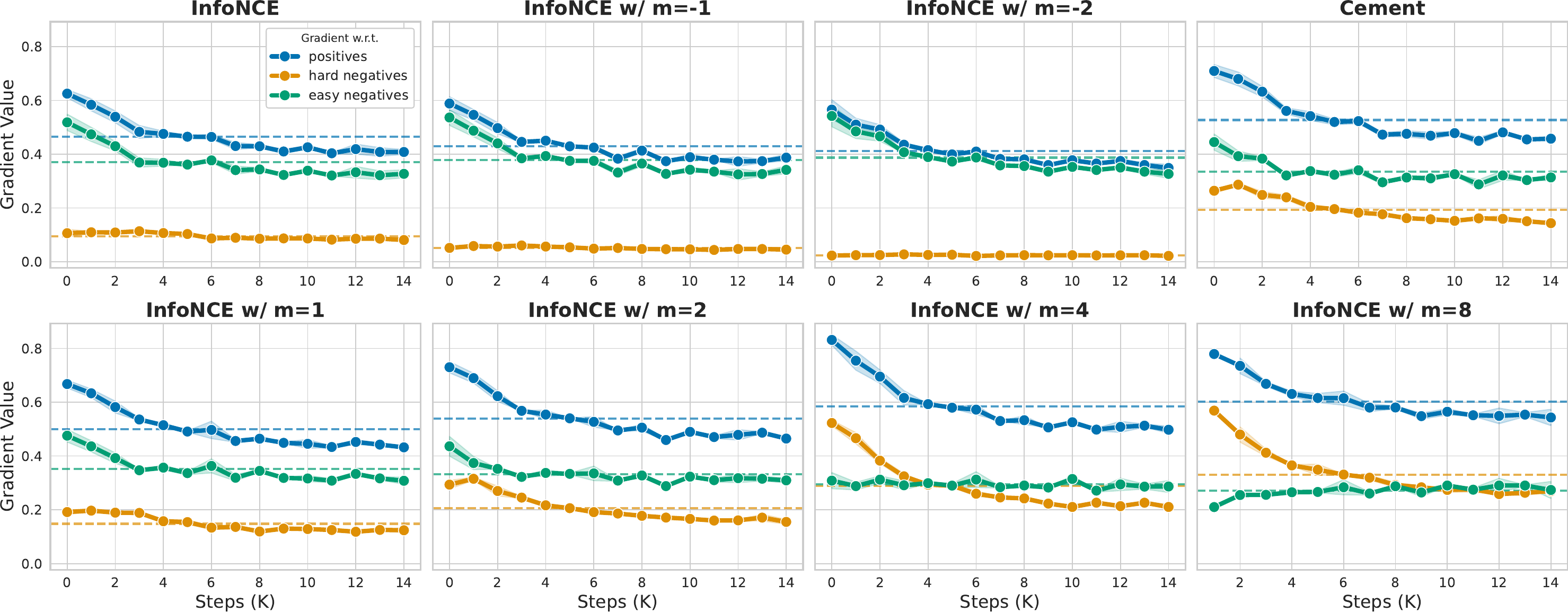}
    \caption{
    Gradient magnitude \wrt~$s_{i,i}$, $s_{i,i'}$, $\sum_{j}s_{i,j}$.
    The dotted lines indicate the average gradient values across the training steps.
    Severe gradient imbalance cause catastrophic forgetting but can be controlled by margin.
    Note that $|\frac{\partial L}{\partial s_{i,i}}s_{i,i}|=|\frac{\partial L}{\partial s_{i,i'}} s_{i,i'}|+|\frac{\partial L}{\partial s_{i,j}}\sum_{j \notin \{i,i'\}}^{2N}s_{i,j}|$.
    }
    \vspace{-1.5em}
    \label{fig:gradient-observation}
\end{figure*}

Gradient magnitudes \wrt~each components are derived to probe the learning dynamics (Details in Appendix~\ref{appendix:gradient-derivation}):
\begin{equation}
    \frac{\partial L_{v \to t}}{\partial s_{i,i}} = -(1 - p_{i,i}),\quad \frac{\partial L_{v \to t}}{\partial s_{i,j}} = p_{i,j}.
\end{equation}
Note that $\sum_{j=1}^{2N}p_{i,j}=1$ always holds due to the softmax operation property.
The gradient magnitude attracting the anchor pair can be rephrased as the sum of negative probabilities:
\begin{equation} \label{eq:gradient-sum}
\underbrace{1 - p_{i,i}}_{\text{Positive Attract}} = \underbrace{p_{i,i'}}_{\substack{\text{Hard Negative Reject}}} + \underbrace{\sum\nolimits_{j \notin \{i,i'\}}^{2N} p_{i,j}.}_{\substack{\text{Easy Negative Reject}}}
\end{equation}
Since contrastive learning requires large batch sizes to form well-distributed latent spaces, the aggregated probability mass of the easy negatives ($\sum_{j\notin \{i,i'\}}p_{i,j}$) substantially grows as the batch size scales in the second term on the right hand side of Eq.~\eqref{eq:gradient-sum}.
Interestingly, the top leftmost plot in Fig.~\ref{fig:gradient-observation} shows that the gradient magnitude of easy negatives has \textit{exploded} and consumed at least 72\% of the total gradient signal even with $N=1024$.
Therefore, this imbalanced penalty towards easy negatives results in InfoNCE to focus on adapting to a new data distribution, rather than learning the subtle difference that invalidates the semantics.
We further hypothesize that balancing the two push forces can address the catastrophic forgetting and enhance the gradient flow of compositional structures.

\begin{wrapfigure}{r}{0.35\linewidth}
    \centering
    \vspace{-1em}
    \includegraphics[width=\linewidth]{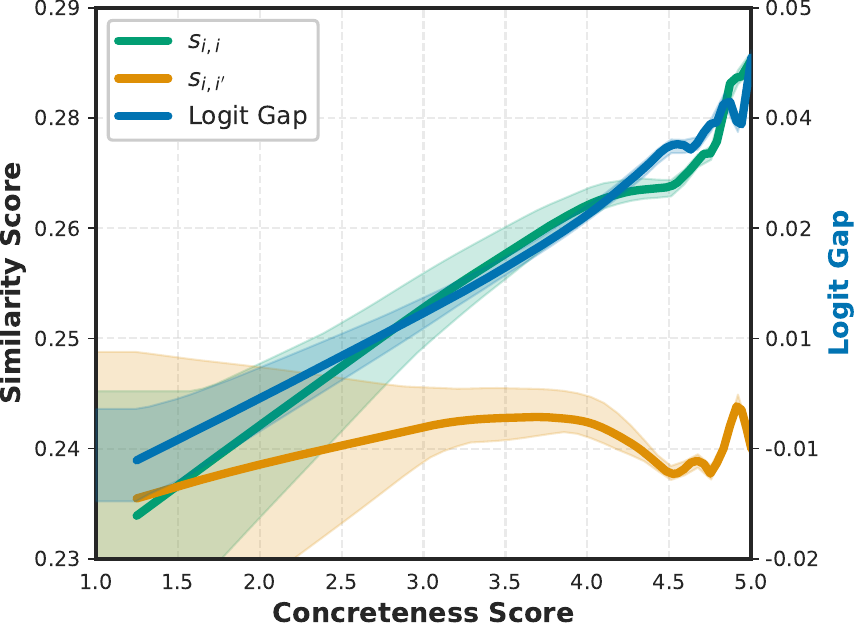}
    \vspace{-1.5em}
    \caption{
    Estimated similarity scores \wrt~$c_i$.
    Curves are fitted with LOESS~\citep{cleveland1979robust}.
    }
    \vspace{-1em}
    \label{fig:concreteness-gap}
\end{wrapfigure}

To mitigate this gradient imbalance problem, a simple yet effective solution is suggested: adding a margin $m$ to the hard negative similarity term in the softmax computation, such that $\tilde{s}_{i,i'} = s_{i,i'} + m$.
This artificially inflates exponential term for the hard negative in the partition function.
Leveraging the prior that Eq.~\eqref{eq:gradient-sum} is a zero-sum, increasing the probability mass of the hard negative forces a proportional reduction in the easy negatives.
This effectively penalizes the gradient bandwidth from the $\sum_{j\notin \{i,i'\}}p_{i,j}$ and reallocate it to the fine-grained compositional semantic boundary.
The second row in Fig.~\ref{fig:gradient-observation} show that the overall ratio of the hard negative signal increases proportional to the scale of $m$.
One may consider removing the easy negatives from the InfoNCE denominator, but removing the inductive bias of discriminating false negatives may trigger serious catastrophic forgetting and unlearn the uniformity~\citep{wang2020understanding, cai2020all}.
This tradeoff leads to another research question: \textit{How can we adaptively determine the margin value?}

Correlation between $s_{i,i}$, $s_{i,i'}$ and $c_i$ are further examined, inspired by the observation that $c_i$ serves as a reliable quality metric for $(v_i',t_i')$.
Hard negatives with lower $c_i$ exhibit smaller $s_{i,i}-s_{i,i'}$, with the gap increasing linearly with $c_i$.
This pattern suggests that high $c_i$ samples encode more informative compositional features, attributed by clearer visual distinction.
In these cases, a larger margin is encouraged to provide a stronger compositional learning signal, whereas noisy signals from visually indistinguishable samples from low $c_i$ should map to lower $m$ to spare gradient capacity for easy negatives.

To counter amplifying undesirable signals, the margin is rectified according to $c_i$ with a bounded smoothing function.
Due to the sampling bias in the data generation process, the distribution
The function has to be bounded by $\left[m_\text{min}, m_\text{max}\right]$ to prevent excessive penalization.
Because of the sampling bias from the data generation process, the distribution of $c_i$ is highly left-skewed.
To address this asymmetry, Fermi-Dirac distribution~\citep{fermi1926sulla, dirac1926theory, kim2021fermi} is adopted to induce a phase transition around $c_i=\tilde{c}$, while the margin is constrained within $\left[m_\text{min}, m_\text{max}\right]$ to avoid excessive penalization.
The adaptive margin is formulated as:
\begin{equation}
    \hat{m}_i = \frac{m_{\max}-m_{\min}}{1+\exp ((\tilde{c}-c_i)/\tau_m)} + m_{\min},
\end{equation}
where $\tau_m$ and $\tilde{c}$ indicate steepness and threshold parameters, respectively.
Let $Z_i^{v \to t}$ denote the margin-injected partition function for a given vision-to-text anchor.
Eq.~\eqref{eq:cement} formalize the \textbf{C}oncreteness-\textbf{e}nhanced \textbf{m}argin \textbf{e}stimation for \textbf{n}egative \textbf{t}raining (\textbf{Cement}) loss.
The model trained with ConcreteBatch is referred to \textbf{S}elective \textbf{l}anguage-\textbf{i}mage \textbf{p}retraining \textbf{f}or \textbf{o}ptimized \textbf{r}epresentation \textbf{m}ining (\textbf{Slipform}).
\begin{equation} \label{eq:cement}
\begin{split}
    L_\text{Cement} &= -\frac{1}{2N} \sum_{i=1}^{2N} \left( \log \frac{\exp(s_{i,i})}{Z_i^{v \to t}} + \log \frac{\exp(s_{i,i})}{Z_i^{t \to v}} \right),\\
    \text{s.t.}\quad Z_i^{v \to t} &= \exp(s_{i,i}) + \exp(s_{i,i'} + \hat{m}_i) + \sum_{j \notin \{i,i'\}}^{2N} \exp(s_{i,j}).
\end{split}
\end{equation}

\section{Experiments}
\paragraph{Experimental Setups}
\begin{figure*}[t]
    \centering
    \begin{subfigure}[t]{0.325\textwidth}
        \centering
        \includegraphics[width=\textwidth]{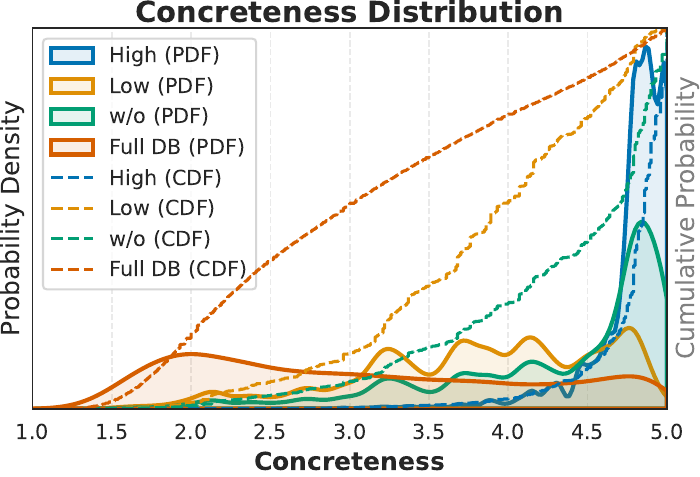}
        \caption{\centering
        $c_i$ distribution by datasets.
        }
        \label{fig:concreteness-distribution}
    \end{subfigure}
    \hfill
    \begin{subfigure}[t]{0.325\textwidth}
        \centering
        \includegraphics[width=\textwidth]{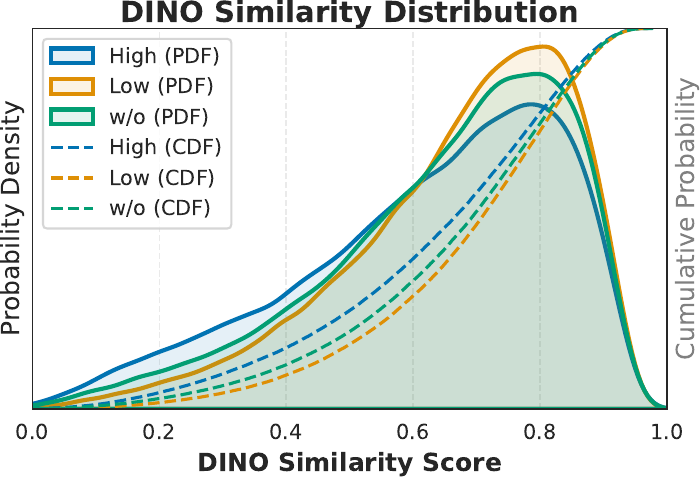}
        \caption{\centering
        DINOScore by datasets, $\mathcal{D}_{hc}$ has lowest similarities.
        }
        \label{fig:dinoscore}
    \end{subfigure}
    \hfill
    \begin{subfigure}[t]{0.325\textwidth}
        \centering
        \includegraphics[width=\textwidth]{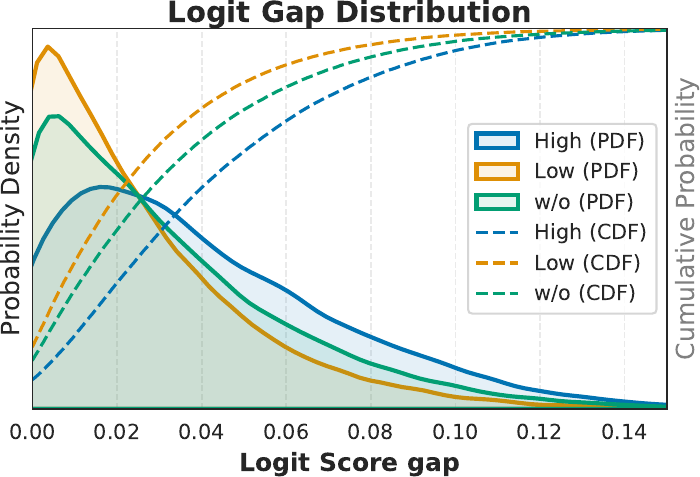}
        \caption{\centering
        Hardness by datasets, attributed by DINOScore.
        }
        \label{fig:logit-gap}
    \end{subfigure}

    \begin{subfigure}[t]{0.325\textwidth}
        \centering
        \includegraphics[width=\textwidth]{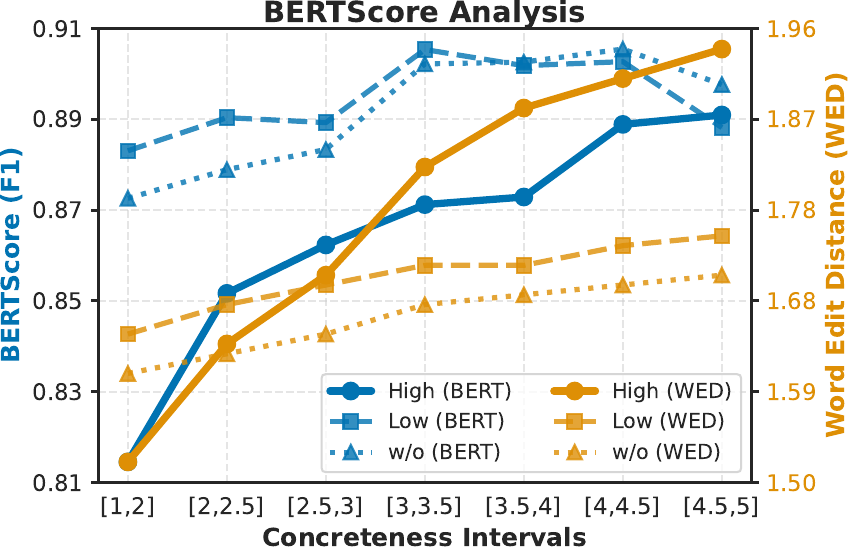}
        \caption{\centering
        BERTScore is correlated with bi-gram frequency.
        }
        \label{fig:bertscore-analysis}
    \end{subfigure}
    \hfill
    \begin{subfigure}[t]{0.325\textwidth}
        \centering
        \includegraphics[width=\textwidth]{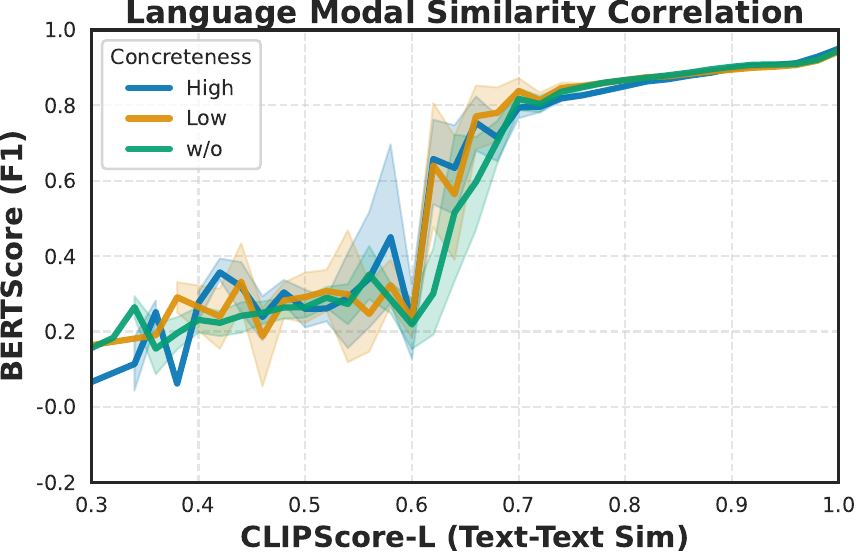}
        \caption{\centering
        Correlation between CLIPScore-L and BERTScore.
        }
        \label{fig:clip-bert-corr}
    \end{subfigure}
    \hfill
    \begin{subfigure}[t]{0.325\textwidth}
        \centering
        \includegraphics[width=\textwidth]{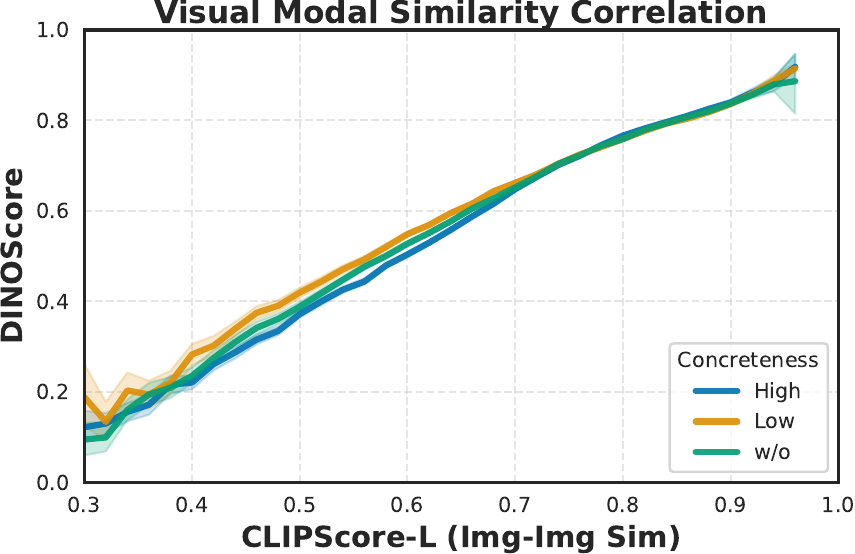}
        \caption{\centering
        Correlation between CLIPScore-V and DINOScore.
        }
        \label{fig:clip-dino-corr}
    \end{subfigure}
    \caption{
    Analysis of dataset distributions and metric correlations regarding concreteness scores, dataset difficulties and intra-modal similarities.
    }
    \label{fig:comprehensive-analysis}
    \vspace{-1.5em}
\end{figure*}

MS-COCO Karapathy train split~\citep{lin2014microsoft, karpathy2015deep} is utilized to generate ConcreteBatch generation.
All experiments are implemented with \texttt{open\_clip} library with an \texttt{openai} pretrained ViT-B-32 backbone.
Training is conducted on a single NVIDIA-H200 GPU with a batch size of 1024.
Adaptive margin hyperparameters are determined as $\left[m_\text{min},m_\text{max}\right]=[-2,2]$, $\tilde{c}=4$, and $\tau_m = 0.15$.
Evaluations are categorized into compositional understanding and general visual representation benchmarks.
Compositional understanding benchmarks assess the cross-modal accuracies using captions with swapped key nouns or subtle perturbations of attributes, relations, and objects.
The selected benchmarks include SugarCrepe~\citep{hsieh2023sugarcrepe}, SugarCrepe++~\citep{dumpala2024sugarcrepe++} and Winoground~\citep{thrush2022winoground}.
General visual representation benchmarks evaluate broader capabilities through downstream classification and retrieval.
General visual representation is evaluated via 10-epoch linear probing in two settings: single-label classification on ImageNet-1k~\citep{russakovsky2015imagenet}, reported with top-1 and top-5 accuracy, and multi-label classification on MS-COCO~\citep{lin2014microsoft}, reported with mean Average Precision (mAP).
The zero-shot retrieval is assessed on Flickr30k~\citep{young2014image} using Recall@K.
VTAB benchmark~\citep{zhai2019large} follows the standard protocol of fitting a logistic regression classifier on each 19 subcategories to the frozen visual features.

\subsection{Dataset Analyses}
\begin{wraptable}{r}{0.35\linewidth}
    \centering
    \vspace{-1em}
    \resizebox{\linewidth}{!}{
        \begin{tabular}{lccc}
            \rowcolor{HeaderBlue}
            \toprule
            \textbf{Metrics} & $\mathcal{D}_{wo}$ & $\mathcal{D}_{lc}$ & $\mathcal{D}_{hc}$ \\
            \midrule
            Mean $c_{i} \uparrow$ & 4.29 & 3.81 & \textbf{4.77} \\
            Median $c_{i} \uparrow$ & 4.63 & 3.89 & \textbf{4.85} \\
            I2T acc. $\uparrow$ & 0.8712 & 0.8357 & \textbf{0.9225} \\
            T2I acc. $\uparrow$ & 0.4433 & 0.4020 & \textbf{0.5087} \\
            \midrule
            BERTScore & 0.8992 & 0.8981 & 0.8907 \\
            CLIP-L & 0.9197 & 0.9279 & 0.9029 \\
            DINOScore & 0.6514 & 0.6707 & 0.6225 \\
            CLIP-V & 0.7168 & 0.7270 & 0.7000 \\
            \#~Unique $\uparrow$ & 11050 & 11034 & \textbf{11190} \\
            \bottomrule
        \end{tabular}
    }
    \vspace{-0.5em}
    \captionof{table}{Statistics summary of datasets by sampling bias.}
    \label{tab:dataset-metrics}
    \vspace{-1.5em}
\end{wraptable}
Detailed investigations of the ConcreteBatch are performed regarding sampling bias, hardness, and intra-modal similarities.
CLIPScores are measured with a PE-Core-L-14-336~\citep{bolya2025perception} pretrained on the MetaCLIP-5.4B dataset~\citep{xu2023demystifying} (total 58B samples seen), DINOv2~\citep{oquab2023dinov2} is employed for DINOScores, and BERTScores~\citep{zhang2019bertscore} is calculated using the default RoBERTa-large~\citep{liu2019roberta} with baseline rescaling.
For notational convenience, let $\mathcal{D}_{hc}$, $\mathcal{D}_{lc}$ and $\mathcal{D}_{wo}$ denote datasets generated by perturbing keywords with high, low, and randomly sampled concreteness, respectively.
First two rows in Tab.~\ref{tab:dataset-metrics} and Fig.~\ref{fig:concreteness-distribution} illustrate highly left-skewed $c_i$ distribution for $\mathcal{D}_{hc}$ compared to $\mathcal{D}_{wo}$, while $\mathcal{D}_{lc}$ exhibits a more uniform distribution.
These results validate the controllability of the proposed data generation pipeline.

The third and fourth rows of Tab.~\ref{tab:dataset-metrics} quantify dataset hardness through cross-modal retrieval accuracies.
As visualized in Fig.~\ref{fig:dinoscore} and Fig.~\ref{fig:logit-gap}, $\mathcal{D}_{lc}$ exhibits the smallest logit gaps and the highest DINOScore, whereas $\mathcal{D}_{hc}$ demonstrates the inverse trend, and $\mathcal{D}_{wo}$ maintains intermediate values for both metrics.
These results validates our findings in Sec~\ref{sec:adaptive-margin} and support the main claim that higher concreteness scores facilitates more distinguishable visual contents in hard negative images.
The model achieves higher T2I and I2T retrieval accuracies on the highly concrete subset since visual semantic shifts in $\mathcal{D}_{hc}$ are more perceptible than those in $\mathcal{D}_{lc}$.
Conversely, the low-concreteness keywords in $\mathcal{D}_{lc}$ produce more imperceptible visual modifications, resulting in tighter logit gaps and increased retrieval difficulty.

Intra-modal semantic similarities are further examined to contextualize these visual findings.
A potential concern arises if a decrease in textual similarity implies a semantic drift so drastic that the generated samples no longer function as valid hard negatives.
This is first addressed by analyzing the linguistic structural shifts in Fig.~\ref{fig:bertscore-analysis}.
While $c_i$ inversely correlates with BERTScore values, this perceived linguistic shift is strongly coupled with an increased mean word edit distance.
This trend is primarily driven by the structural nature of highly concrete entities, which frequently manifest as bi-grams (\eg, ``coffee mug'' or ``fire hydrant'').
Consequently, the observed BERTScore variance is a byproduct of manipulating \textit{semantically dense, multi-word entities rather than a loss of semantic or grammatical coherence}.
To ensure these structural linguistic modifications translate consistently into the visual domain, visiolinguistic correlations are examined in Figs.~\ref{fig:clip-dino-corr} and \ref{fig:clip-bert-corr}.
These correlations explain low intra-modal similarity metrics of $\mathcal{D}_{hc}$ are not mainly stemmed from visiolinguistic misalignment or validity of the hard negatives.

\subsection{Gradient of InfoNCE}
\begin{wrapfigure}{r}{0.4\linewidth}
    \centering
    \vspace{-5em}
    \includegraphics[width=\linewidth]{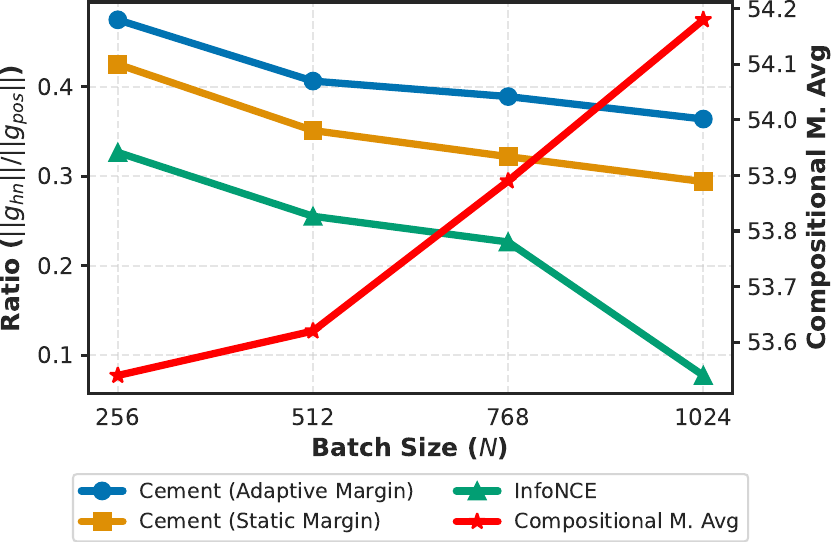}
    \vspace{-1.5em}
    \caption{
    Gradient imbalance gets worse as batch size grows, but reducing the batch size degrades performance.
    }
    \vspace{-2em}
    \label{fig:batchsize}
\end{wrapfigure}

The theoretical gradient magnitude disparity raised in Sec.~\ref{sec:adaptive-margin} is empirically validated by tracking the average gradient allocations during training.
Fig.~\ref{fig:batchsize} plots the ratio of the gradient magnitude towards hard negative relative to the total positive pull gradient across the varying batch size.
A clear downward trend is observed as the training batch size grows, InfoNCE falls below 20\% at $N=1024$.

A naive solution to this imbalance is to reduce the batch size, as $N=256$ recovers the InfoNCE gradient ratio to a healthier level above 0.32.
Contrastive learning intrinsically relies on a massive number of negative samples to approximate the data distribution and enforce latent space uniformity. 
Empirical evaluation demonstrates that shrinking the batch size from 1024 to 256 incurs a 2.16\% relative performance loss in general visual representation alongside a 1.18\% relative degradation in compositional understanding. 
Therefore, sacrificing batch size to resolve the hard negative signal is not a viable strategy.
Margin addition is a better solution to penalize the gradient discrepancy and retain the obtainable knowledge from large batch size.

\subsection{Quantitative Results}
The effectiveness of the ConcreteBatch and Cement loss are empirically evaluated.
Tab.~\ref{tab:compositional-benchmark} and Tab.~\ref{tab:visual-representation} report subcategory scores and macro-average of compositional understanding benchmarks and general visual representation tasks, respectively.
Inverse margin indicates the model trained with $\hat{m}_i=(m_{\max}-m_{\min})/(1+\exp((c_i-\tilde{c})/\tau_m))+m_{\min}$ and static margin of 1 is determined for comparison.
The proposed \model~achieves state-of-the-art performance across the evaluated compositional understanding benchmarks, demonstrating 13.13\% relative gain in the macro average compared to the standard CLIP baseline.
In addition, a consistent performance trend of datasets with different $c_i$ sampling strategies validates the core hypothesis regarding the effectiveness of utilizing highly concrete hard negatives for training.

Furthermore, sustained improvements from Cement loss against InfoNCE empirically confirm that addressing the gradient magnitude imbalance allows models to focus on penalizing hard negatives and utilize them more effectively.
As expected, applying a static margin yields substantial performance gains, while inverse margin degrades the compositional understanding by encouraging to prioritize easy negatives penalties.
Meanwhile, a trade-off is observed between compositional understanding and general visual representation performance while employing the Cement loss and $\mathcal{D}_{hc}$.
The observed performance pattern suggests that optimizing for fine-grained compositional discrimination introduces a meaningful but natural tension with broader visual representation objectives.
In particular, the adaptive margin appears to successfully steer learning toward harder compositional cases, while the challenging distribution of $\mathcal{D}_{hc}$ further reinforces this behavior.
Nevertheless, \model~maintains general visual understanding performance that is comparable to other methods, while achieving our primary objective of improving compositional understanding.
Exploring how to further enhance both aspects simultaneously through other training factors remains an important direction for future work.

\begin{table*}[t]
    \vspace{-0.5em}
    \begin{adjustbox}{max width=\textwidth, center}
    \begin{tabular}{lccccccc}
        \toprule
        \rowcolor{HeaderBlue}
        \textbf{Model} & \textbf{SugarCrepe$^\uparrow$} & \textbf{SCpp-I2T$^\uparrow$} & \textbf{SCpp-T2I$^\uparrow$} & \textbf{WG-txt$^\uparrow$} & \textbf{WG-img$^\uparrow$} & \textbf{WG-grp$^\uparrow$} & \textbf{M.Avg$^\uparrow$} \\
        \midrule
        CLIP & 75.38 & 59.41 & 46.84 & 29.00 & 9.00 & 7.50 & 47.89 \\
        NegCLIP & 82.50 & 62.75 & 58.15 & 30.50 & 11.00 & 8.00 & 53.15 \\
        TSVLC & 79.07 & 55.30 & 50.05 & 21.25 & 8.50 & 4.75 & 47.75 \\
        TSVLC+ & 72.38 & 52.49 & 49.00 & 24.00 & 8.00 & 4.75 & 45.13 \\
        TripletCLIP & 82.47 & 58.43 & 53.00 & 25.75 & 11.00 & 7.00 & 50.92 \\
        CE-CLIP & 85.25 & 47.66 & 53.18 & 18.50 & 11.25 & 5.00 & 49.08 \\
        DeGLA & 89.23 & 62.84 & 54.94 & 24.50 & 10.00 & 7.25 & 54.01 \\
        \midrule
        $\mathcal{D}_{lc}$ + InfoNCE & 78.01 & 63.38 & 60.59 & 31.25 & 9.25 & 7.75 & 52.03 \\
        $\mathcal{D}_{wo}$ + InfoNCE & 78.28 & 63.42 & 60.26 & 32.25 & 9.75 & 8.00 & 52.26 \\
        $\mathcal{D}_{hc}$ + InfoNCE & 81.70 & 65.09 & 60.68 & 31.75 & 10.75 & 8.75 & 53.89 \\
        $\mathcal{D}_{lc}$ + Cement & 82.85 & 65.62 & 61.05 & 29.00 & 9.75 & 6.25 & 53.72 \\
        $\mathcal{D}_{wo}$ + Cement & 81.76 & 64.93 & 60.79 & 30.25 & 12.00 & 8.25 & 53.82 \\
        Inverse Margin & 80.68 & 64.41 & 60.12 & 30.75 & 9.75 & 7.00 & 52.92 \\
        Static Margin & 82.31 & 65.81 & 60.14 & 32.50 & 9.75 & 8.50 & 54.07 \\
        \rowcolor{gray!10} Slipform & 83.00 & 66.24 & 60.21 & 31.00 & 10.25 & 7.75 & 54.18 \\
        \bottomrule
    \end{tabular}
  \end{adjustbox}
  \vspace{-0.5em}
  \caption{
  Compositional understanding benchmark results, M.Avg indicates the macro accuracy.
  Results validate the effectiveness of the proposed dataset and objective.
  }
  \vspace{-0.5em}
  \label{tab:compositional-benchmark}
\end{table*}
\begin{table}[t]
    \centering
    \begin{adjustbox}{max width=\textwidth, center}
    \begin{tabular}{lccccccccc}
        \toprule
        \rowcolor{HeaderBlue}
        & \textbf{COCO} & \multicolumn{2}{c}{\textbf{ImageNet-1k}} & \multicolumn{4}{c}{\textbf{Flickr30k}} & \textbf{VTAB} & \\
        \arrayrulecolor{HeaderBlue}\specialrule{6pt}{0pt}{-6pt}\arrayrulecolor{black}
        \cmidrule(lr){2-2} \cmidrule(lr){3-4} \cmidrule(lr){5-8} \cmidrule(lr){9-9} 
        \rowcolor{HeaderBlue}
        \multirow{-2.4}{*}{\textbf{Method}} & \textbf{mAP$^\uparrow$} & \textbf{top1$^\uparrow$} & \textbf{top5$^\uparrow$} & \textbf{T2I-R@1$^\uparrow$} & \textbf{T2I-R@5$^\uparrow$} & \textbf{I2T-R@1$^\uparrow$} & \textbf{I2T-R@5$^\uparrow$} & \textbf{Mean$^\uparrow$} & \multirow{-2.4}{*}{\textbf{M.Avg$^\uparrow$}} \\
        \midrule
        CLIP & 24.43 & 43.12 & 70.26 & 21.15 & 40.51 & 37.04 & 61.00 & 33.08 & 38.53 \\
        NegCLIP & 26.57 & 43.37 & 71.04 & 28.84 & 51.14 & 42.18 & 65.95 & 34.11 & 41.23 \\
        TSVLC & 25.53 & 43.98 & 71.08 & 23.55 & 43.84 & 23.82 & 45.01 & 33.77 & 37.72 \\
        TSVLC+ & 25.48 & 43.89 & 71.15 & 23.94 & 44.36 & 26.40 & 48.95 & 33.83 & 38.19 \\
        TripletCLIP & 25.08 & 40.70 & 67.84 & 19.25 & 37.59 & 16.67 & 35.77 & 32.11 & 34.69 \\
        CE-CLIP & 24.74 & 46.02 & 74.91 & 26.19 & 47.34 & 22.71 & 45.28 & 39.33 & 39.98 \\
        DeGLA & 25.81 & 44.24 & 71.60 & 30.13 & 52.70 & 37.04 & 60.95 & 34.26 & 40.80 \\
        \midrule
        $\mathcal{D}_{lc}$ + InfoNCE & 26.42 & 43.96 & 71.34 & 30.29 & 52.70 & 48.02 & 71.09 & 34.15 & 42.19 \\
        $\mathcal{D}_{wo}$ + InfoNCE & 26.44 & 44.00 & 71.45 & 30.35 & 52.67 & 47.96 & 71.08 & 34.15 & 42.18 \\
        $\mathcal{D}_{hc}$ + InfoNCE & 26.21 & 42.41 & 69.68 & 29.43 & 51.46 & 45.09 & 68.45 & 33.18 & 41.01 \\
        $\mathcal{D}_{lc}$ + Cement & 26.00 & 42.27 & 69.72 & 29.17 & 51.42 & 42.92 & 66.78 & 33.42 & 40.75 \\
        $\mathcal{D}_{wo}$ + Cement & 26.24 & 42.28 & 69.52 & 29.79 & 52.05 & 46.35 & 69.68 & 33.15 & 41.19 \\
        Inverse Margin & 26.27 & 42.46 & 69.81 & 29.82 & 51.90 & 46.55 & 69.63 & 33.33 & 41.30 \\
        Static Margin & 26.19 & 42.67 & 70.09 & 28.97 & 50.93 & 43.56 & 66.96 & 33.45 & 40.91 \\
        \rowcolor{gray!10} Slipform & 26.16 & 43.11 & 70.57 & 28.44 & 50.42 & 42.05 & 65.58 & 33.64 & 40.82 \\
        \bottomrule
    \end{tabular}
    \end{adjustbox}
    \vspace{-0.5em}
    \caption{
    General visual representation benchmark results, M.Avg indicates the macro accuracy.
    A tradeoff between compositional understanding is observed.
    }
    \vspace{-0.5em}
    \label{tab:visual-representation}
\end{table}
\section{Discussion}

\paragraph{Limitations and Future Work}
Although this study successfully integrates the heuristics of concreteness-based perturbation into data generation, several challenges remain unresolved.
First, standard concreteness scores also account for non-visual perceptibilities, including tactile, auditory, olfactory, and gustatory senses.
Enhancing existing visual-specific concreteness measures~\citep{hessel2018quantifying} and incorporating them into the \datapipe~framework may yield a more precise approximation of latent visual concreteness.
Second, further investigation is required to mitigate the observed performance trade-off between compositional understanding and general visual representation.
Finally, future research will explore the impact and correlations of these generated datasets within downstream tasks~\citep{im2023deep}, video applications~\citep{ali2024harnessing}, and their utility as submodules~\citep{lai2024revisit, li2025vidhalluc} to further evaluate the compositional reasoning and generation.
\section{Related Work}

\paragraph{Improving Cross-Modal Pretraining}
Recent studies explored how to address the limitations inherent in contrastive cross-modal pretraining objectives and datasets.
SigLIP~\citep{zhai2023sigmoid, tschannen2025siglip} improves training efficiency by substituting the softmax operation with a binary cross-entropy formulation.
Additionally, DCL~\citep{yeh2022decoupled} mitigates the gradient vanishing problem and addresses the push-pull imbalance inherent within the InfoNCE objective~\citep{oord2018representation}.
To improve dataset quality, researchers filter misaligned web-crawled pairs using static~\citep{hessel2021clipscore} and dynamic~\citep{zhao2025differential} thresholding techniques.
Other approaches perform recaptioning to replace noisy text descriptions~\citep{lai2024veclip} and enhance caption granularity~\citep{doveh2023dense, lai2024revisit}.
Standard multimodal contrastive optimization requires exceptionally large training batch sizes to achieve peak performance.
Consequently, alternative training strategies explore smaller batch sizes~\citep{yuan2022provable} and memory-efficient frameworks~\citep{chen2023disco}. 
Furthermore, web-scale image-caption datasets frequently lack complex linguistic phenomena.
These missing elements include negations~\citep{alhamoud2025vision, park2025know}, precise attributes~\citep{paiss2023teaching, li2024nemo}, and reliable spatial prepositions~\citep{kamath2023s, wang2025spatial457}.
The absence of these subtle semantic differences in the training batch creates an inductive bias that leads to near-random performance on compositional understanding benchmarks~\citep{thrush2022winoground, ma2023crepe, hsieh2023sugarcrepe, awal2024vismin}.
This bias ultimately restricts the efficacy of vision-language models when deployed as neural modules for complex reasoning~\citep{ke2025explain} and generation tasks~\citep{singh2025chimera}.

\paragraph{Hard Negative Mining}
Hard negative mining focuses on automating and scaling hard negative generation.
These efforts primarily investigate methods for perturbing semantics and introducing heuristics to mitigate modality bias~\citep{chung2025mass}.
Recent approaches utilize both visual and textual generative models to automate hard negative mining and finetune to correct the bias~\citep{fan2023improving, patel2024tripletclip}.
Moreover, graph priors serve to improve structured semantic representations~\citep{singh2023coarse, huang2024structure}.
Zhang~\etal~\citep{zhang2024contrasting} proposed an intra-modal similarity penalty between positive and hard negative pairs.
HardPositive~\citep{kamath2024hard} observed that negative training can introduce oversensitivity to perturbations and induce structural brittleness.
However, current literature focuses on the mechanics of generation but fails to establish a systematic criterion for determining exactly \textit{which} semantic element to perturb.

\section{Conclusion}

This research established a concreteness-aware approach to hard negative mining for VLMs. Lexical concreteness was identified as the primary determinant of negative sample quality. Analysis revealed that modifying concrete concepts systematically produces larger visual discrepancies and informative contrastive supervision. To leverage this finding, the ConcretePlant was designed to control the factor, and generate perceptually grounded negative pairs. Furthermore, the InfoNCE was shown to suffer from a gradient imbalance. The proposed loss resolves this degradation by employing a concreteness-conditioned adaptive margin to retain the hard negative signals. Empirical evaluations confirmed that the resulting model achieves state-of-the-art accuracy on compositional understanding. 
The findings highlight the importance of aligning data generation strategies with objective design to enhance multimodal compositional reasoning.

\bibliography{reference}
\bibliographystyle{unsrtnat}

\clearpage
\setcounter{page}{1}
\appendix
\section*{\Large Concrete Jungle: Towards Concreteness Paved Contrastive Negative Mining for Compositional Understanding}

\section{More Details on ConcretePlant}
\subsection{Concreteness-Aware Data Generation}
\label{appendix:method-detail}
To utilize the finest granularity~\citep{doveh2023dense, lai2024revisit} among the available captions annotated to images in the dataset~\citep{lin2014microsoft}, we first select the caption with the largest token count as the anchor caption, under the assumption that longer captions typically convey richer compositional structure.
Then we parse each anchor caption with SpaCy~\citep{honnibal2017spacy} to obtain tokenization, lemmatization, part-of-speech tags, and dependency structure. This linguistic parse enables us to (i) enumerate candidate perturbation targets and (ii) determine whether captions are perturbed with correct grammar.

Given a set of anchor captions, each caption is decomposed into a set of $n$ key entities $t_i = [t_i^1, ..., t_i^n]$, and lookup the corresponding concreteness rating~\citep{brysbaert2014concreteness} to build top-$k$ filtered score set $C_i = [c_i^1, ..., c_i^k]$ for each $i$.
\footnote{POS taggings other than
$\{$\texttt{NOUN}, \texttt{PROPN}, \texttt{ADJ}, \texttt{NUM}, \texttt{ADP}, \texttt{VERB}$\}$ are filterd to focus the perturbation on semantically meaningful, perceptually grounded units while excluding function words (\eg, determiners and conjunctions) that rarely correspond to perceptible concepts.}
Top-$k$ sampling from $\text{softmax}(C_i)$ is performed to preserve the bias toward highly perceptible targets while avoiding mode collapse that would arise from always perturbing the single most concrete token, thereby improving coverage and diversity across the generated negatives.
We determine the compositional type for each $i$ from well-established categories in the literature: \texttt{attribute}, \texttt{relation}, and \texttt{object} (ARO)~\citep{yuksekgonul2022and}.
Due to the relative sparseness of relational descriptions in $\mathcal{D}$, we prioritize the \texttt{relation} type if it exists.
We apply deficit-based quota sampling to balance the \texttt{attribute} and \texttt{object} class.

\paragraph{Balancing Compositional Class}
Attribute is enabled when the caption contains a well-formed attribute handle (\eg, adjectival or numeric modifiers, or state-like verb attributes) that can be substituted without invalidating the grammar.
Object is enabled when a concrete noun/proper noun entity can be replaced directly.
Relation is enabled only when the caption exhibits spatial or interaction structure, detected via (i) explicit multi-word spatial phrases and/or (ii) spatial prepositions with a nominal object in the dependency tree.
Additionally, interaction-style relations are identified through transitive verbs with subject-object structure.

\subsection{Prompt}
\label{appendix:prompt}
The following is the vertabim of the prompt.
Different in-context examples are used for different compositional categories.
\begin{tcolorbox}[enhanced, frame hidden, borderline west={2pt}{0pt}{white!75!black}, colback=white, breakable]
\small
**Task**: You are an expert data engineer. Given a conceptual caption, your goal is to generate a new caption that is physically and grammatically correct but factually mismatched to the original image.

**Instruction**:
1. You will be provided with an original caption and a target keyword present in the caption.
2. Locate the keyword within the caption and replace it with a new word or phrase that creates a clear visual conflict.
3. Keep the sentence structure, grammar, and all other words identical to the original.
4. The output must be physically sensible (e.g., "A hat wearing a man" is NOT allowed).
5. ONLY output the new caption, do not include reasons for the output.

**Examples**:
\{\}

Input: \{\}

Keyword: \{\}

Output:
\end{tcolorbox}
\begin{tcolorbox}[enhanced, frame hidden, borderline west={2pt}{0pt}{white!75!black}, colback=white, breakable]
\small
\# Attribute category

Input: A person wearing a red helmet,
Keyword: red,
Output: A person wearing a blue helmet.
Input: A wooden boat floating on the water,
Keyword: wooden,
Output: A plastic boat floating on the water.
Input: The cloudy sky over the city,
Keyword: cloudy,
Output: The clear sky over the city.

\# Relation category

Input: A cat sitting under a table,
Keyword: under,
Output: A cat sitting on a table.
Input: A child standing in front of the tree,
Keyword: front,
Output: A child standing behind the tree.
Input: A lamp placed to the left of the sofa,
Keyword: left,
Output: A lamp placed to the right of the sofa.

\# Object category

Input: A dog running in the park,
Keyword: dog,
Output: A cat running in the park.
Input: A woman holding a coffee cup,
Keyword: coffee cup,
Output: A woman holding a laptop.
Input: A car driving on the highway,
Keyword: highway,
Output: A car driving on the grass.

\# Without considering concreteness

Input: A person wearing a red helmet,
Keyword: red,
Output: A person wearing a blue helmet.
Input: A child standing in front of the tree,
Keyword: front,
Output: A child standing behind the tree.
Input: A car driving on the highway,
Keyword: highway,
Output: A car driving on the grass.
\end{tcolorbox}

\section{Gradient Derivation}
\label{appendix:gradient-derivation}
To derive the derivative of the InfoNCE loss function $L_\text{InfoNCE}$ with respect to the similarity variable $s_{i,i}$, consider the loss for a single sample index $i$ and its corresponding hard negative index $i'$ in the training batch.

\begin{equation}
L_\text{InfoNCE} = -\frac{1}{2N} \sum_{i=1}^{2N}\log \left( \frac{\exp(s_{i,i})}{\exp(s_{i,i}) + \exp(s_{i,i'}) + \sum_{j \notin \{i,i'\}}^{2N} \exp(s_{i,j})} \right).
\end{equation}
Let the denominator be denoted as $Z$:
\begin{equation}
Z = \exp(s_{i,i}) + \exp(s_{i,i'}) + \sum_{j \notin \{i,i'\}}^{2N} \exp(s_{i,j}).
\end{equation}
Then, the loss can be reformulated as:
\begin{equation}
L_\text{InfoNCE} = -\frac{1}{2N}\sum_{i=1}^{2N}\left( \log(\exp(s_{i,i})) - \log(Z) \right) = -\frac{1}{2N}\sum_{i=1}^{2N}(s_{i,i} - \log(Z)).
\end{equation}
Now, the partial derivative of $L_\text{InfoNCE}$ with respect to $s_{i,i}$ is:
\begin{equation}
\frac{\partial L_\text{InfoNCE}}{\partial s_{i,i}} = \frac{\partial}{\partial s_{i,i}} \left( \log(Z) - s_{i,i} \right) = \frac{\partial}{\partial s_{i,i}} \log(Z) - 1.
\end{equation}
Applying the chain rule to the first term:
\begin{equation}
\frac{\partial}{\partial s_{i,i}} \log(Z) = \frac{1}{Z} \cdot \frac{\partial Z}{\partial s_{i,i}} = \frac{\exp(s_{i,i})}{Z}.
\end{equation}
While substituting $\frac{\partial Z}{\partial s_{i,i}}$ back into the expression, the term $\frac{\exp(s_{i,i})}{Z}$ represents the predicted probability $p_{i,i}$ that the pair $(i, i)$ is the positive match among the candidates. Thus, the derivative formulates as:
\begin{equation}
\frac{\partial L_\text{InfoNCE}}{\partial s_{i,i}} = \frac{\exp(s_{i,i})}{Z} - 1 = -(1-p_{i,i}).
\end{equation}

In case of $\frac{\partial L_\text{InfoNCE}}{\partial s_{i,j}}$,
\begin{equation}
\frac{\partial L_\text{InfoNCE}}{\partial s_{i,j}} = \frac{\partial}{\partial s_{i,j}} \left( \log(Z) - s_{i,i} \right) = p_{i,j}.
\end{equation}
This derivation can be generalized to $\frac{\partial L_\text{InfoNCE}}{\partial s_{i,i'}}=p_{i,i'}$.

\section{Qualitative Samples by Datasets}
\begin{figure*}[t]
    \centering
    \includegraphics[width=\linewidth]{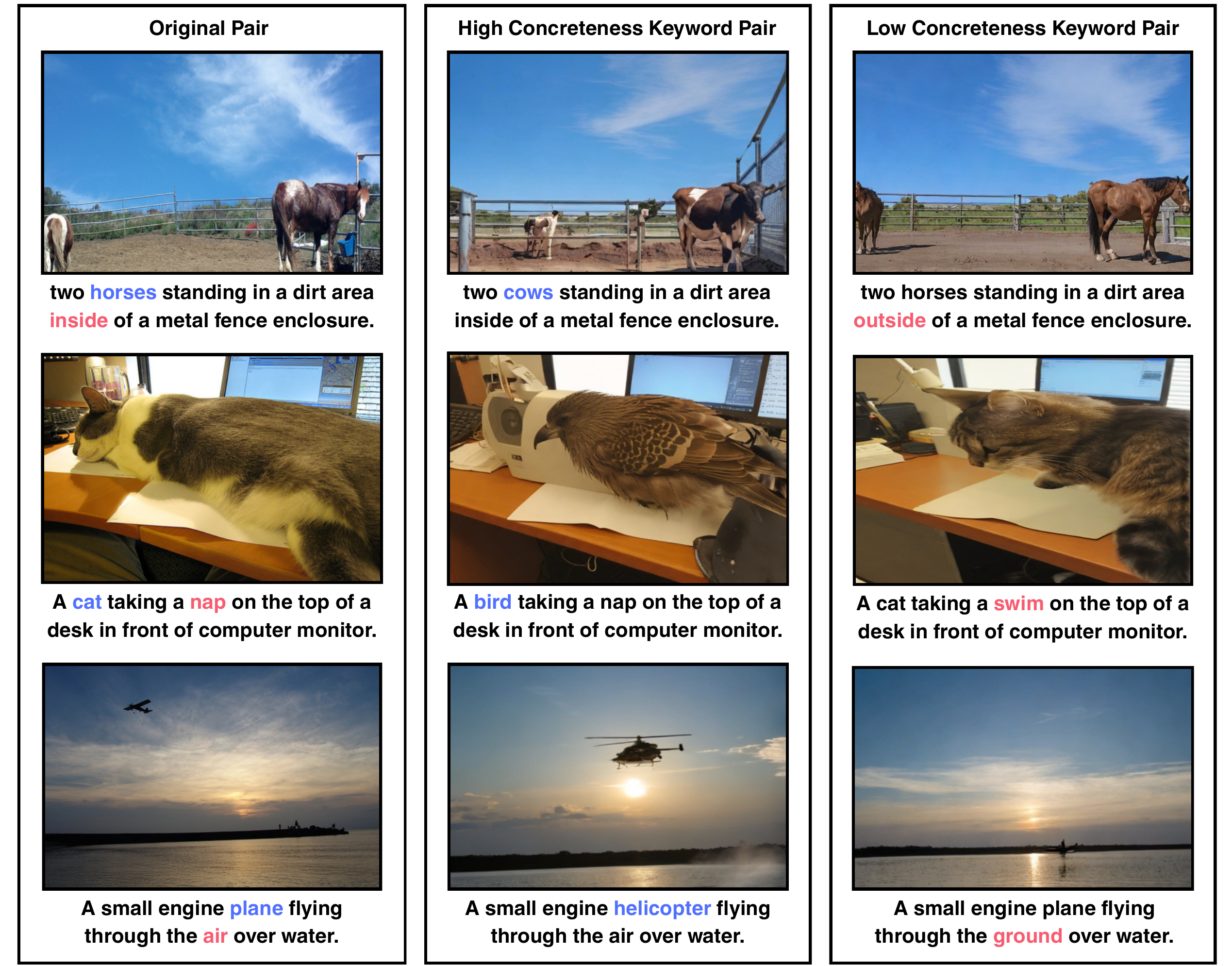}
    \vspace{-1em}
    \caption{
        Qualitative examples of high- and low-concreteness perturbations.
    }
    \vspace{-0.5em}
    \label{fig:cj-qual}
\end{figure*}

To further illustrate the role of concreteness in hard negative generation, Fig.~\ref{fig:cj-qual} provides additional qualitative examples. These comparisons illustrate how perturbing keywords with different levels of concreteness influences the resulting negative pairs. Across the rows, perturbing a highly concrete keyword leads to larger visual or structural changes, while perturbing a lower-concreteness keyword usually yields a subtler modification that preserves more of the original scene. Blue highlights denote high-concreteness edits, and red highlights denote low-concreteness edits.

\end{document}